\documentclass[12pt]{article}

\usepackage[letterpaper,margin=1.25in]{geometry}
\usepackage[utf8]{inputenc}
\usepackage{mathptmx}

\usepackage[tiny]{titlesec}

\usepackage{natbib}
\usepackage{multicol}
\usepackage{float}
\usepackage{graphicx,gb4e} 

\noautomath
\usepackage[utf8]{inputenc}
\usepackage[T1]{fontenc}
\usepackage{longtable} 
\usepackage{hyperref}
\hypersetup{%
	colorlinks,%
	urlcolor=teal,%
	linkcolor=black,%
	citecolor=black}

\usepackage{tipa}
\usepackage{tikz}
\usepackage[linguistics]{forest}
\usetikzlibrary{arrows.meta}

\usepackage{stmaryrd}
\usepackage{gb4e}

\begin{document}
\noindent \textbf{\large A Paradigm Gap in Urdu} \\

\noindent\textsc{Farah Adeeba}, \textit{University of Massachusetts Amherst and University of Engineering \& Technology, Lahore} \\
\textsc{Rajesh Bhatt}, \textit{University of Massachusetts Amherst} \\
\\
\textsc{Abstract}\vspace{-7pt}
\begin{quote}

In this paper, we document a paradigm gap in the combinatorial possibilities of verbs and aspect in Urdu: the perfective form of the -yā: kar construction (e.g., ro-yā: ki: ‘cry-Pfv do.Pfv’) is sharply ungrammatical in modern Urdu and Hindi, despite being freely attested in 19th-century literature. We investigate this diachronic shift through historical text analysis, a large-scale corpus study—which confirms the stark absence of perfective forms—and subjective evaluation tasks with native speakers, who judge perfective examples as highly unnatural. We argue that this gap arose from a fundamental morphosyntactic conflict: the construction's requirement for a nominative subject and an invariant participle clashes with the core grammatical rule that transitive perfectives assign ergative case. This conflict rendered the perfective form unstable, and its functional replacement by other constructions allowed the gap to become entrenched in the modern grammar.

\end{quote}

\section{Introduction}

Human languages are dynamic systems that continually evolve, resulting in the emergence, change, and sometimes complete disappearance of morphological and grammatical structures. Within this diachronic landscape, the phenomenon of \textit{paradigm gaps}—systematic absences of expected word forms or constructions—presents a particularly intriguing puzzle for linguistic theory (\citealt{albright2003}, \citealt{Sims2006} , \citealt{BermelKnittl}). Such gaps challenge models of language production, acquisition, and change, as they represent a failure of the grammar to generate a logically possible form. A notable example is provided by \citet{ileri2018}, who identify a gap in the third person plural paradigm of desiderative constructions in Turkish, a pattern which speakers consistently avoid or deem unacceptable.

This paper investigate the existence of a gap in the combinatorial possibilities of a verb and aspect in Urdu. In general, in Urdu all verbs combine with the full range of aspectual/tense options i.e. given an unfamiliar verb, an Urdu speaker can use it in the perfective, the imperfective, the subjunctive, the future and so on. Given this, it comes as a surprise that the \textit{-ya: kar `-Pfv do’} construction cannot combine with the perfective.

\begin{exe}
\ex
\gll
Vo pehrõ ro-ya: kar-ta: tha:\\
Dem hours cry-Pfv do-Impfv.MSg be.Pst.Msg\\
\glt
He used to cry for hours.
\ex
\gll
*Vo pehrõ ro-ya: kiya:\\
  Dem hours cry-Pfv do.Pfv.MSg\\
\glt
Intended: He cried for hours.

\end{exe}

The puzzle deepens with the finding that this gap is a historical innovation. Through analysis of 19th-century Urdu literature, we show that the perfective `-ya: kar` construction was not only possible but actively used (3).
\begin{exe}
    
    \ex
    \gll Vo raat bhar ro-ya: ki:     (Umrao Jaan Ada, Mir Hadi `Ruswa', 1899)
    \\ Dem night cry-Pfv do-Pfv.F\\ 
    \glt `She cried/kept crying all night.'
    
\end{exe}

This diachronic shift raises two central research questions:
\begin{enumerate}
    \item What factors led to the loss of the perfective \textit{'-ya: kar'} form in modern Urdu?
    \item Why has this specific gap persisted, and how is it learned by speakers, given that it does not create an expressivity problem?
\end{enumerate}

We propose that the gap is motivated by a fundamental \textit{morphosyntactic conflict}. The \textit{'-ya: kar'} construction exhibits two invariant properties: the subject appears in the nominative case, and the participial verb `-ya:` does not agree with the subject. This configuration clashes with a core rule of Urdu grammar: transitive verbs in the perfective aspect typically assign ergative case to their subject. We propose that this conflict rendered the perfective form unstable, leading speakers to avoid it over time. Its semantic function was seamlessly absorbed by other available forms, such as the simple perfective or the `-ta: rah-` construction, thus preventing an ineffability problem and allowing the gap to become entrenched.

To support this argument, we present evidence from multiple sources: a qualitative analysis of historical texts establishing the construction's prior existence, a quantitative corpus study demonstrating its stark absence in modern Urdu, and native speaker judgment tasks confirming its contemporary ungrammaticality.

The paper is structured as follows: Section 2 outlines the two case studies of lost generalizations in Urdu, focusing on the \textit{'ca:-'} and \textit{'-ya: kar'} construction. Section 3 details the syntactic and semantic properties of \textit{'-ya: kar'} and its historical attestation. Section 4 introduces the -ta: rah- construction, which shows no such gap. Section 5 presents our analysis of the case-based motivation for the gap. Section 6 discusses  evaluation of \textit{-Ya: Kar} construction by native speakers. Section 7 concludes. This case study offers a clear model for how paradigm gaps can emerge and stabilize diachronically, driven by syntactic conflict and resolved by competition from existing grammatical forms.

\section{Loss of generalization}
This section examines how a once broadly applicable grammatical construction becomes restricted in its usage over time. A general pattern loses its ability to combine with a wider range of linguistic elements (like verbs or aspects), becoming specialized or even ungrammatical in previously acceptable contexts. This process of reduced applicability demonstrates that language change involves not only innovation but also the narrowing and specialization of existing grammatical structures.

\subsection{Loss of a subcategorization frame}
The verb {\em ca:h-} 'want' provides an example of semantic and syntactic change over time in Hindi-Urdu. The following examples, attested up until the early 20th century, illustrate an earlier usage of {\em ca:h-} where it conveyed an 'about to' or 'imminent' meaning when combined with a perfective verb form:

\begin{exe}
    \ex
    \begin{xlist}
    \ex
    \gll gha\d{r}i: 12 baj-a: ca:h-ti: hai\\
    clock.F  12 ring-Pfv want-Impfv.F be.Prs.Sg\\
    \glt `The clock is about to strike 12.'

    \ex \citep{small1895grammar}
        \gll vo kal aya: ca:h-te hain  \\
        Dem tomorrow come-Pfv want-Impfv.M be.Prs.Pl\\
        \glt 'They will come tomorrow.'    
    \ex
    \gll asar ki: nama:z=ka: waqt hu-a: ca:h-ta: hai\\
    Asar Gen.f prayer=Gen time be-Pfv want-Impfv.M.Sg be.Prs.Sg\\
    \glt `It is about to be time for the Asar prayer.'
    \end{xlist}
\end{exe}

\begin{exe}
    \ex
    \begin{xlist}
    \ex infinitive:
    \gll vo dubai ja:-na: ca:h-ti: hai\\
    Dem Dubai go-Inf want-Impfv.F be.Prs.Sg\\
    \glt `She wants to go to Dubai'
    \ex subjunctive:
    \gll vo  ca:h-ti: hai ki mEN dubai ja:-\~{u}\\
    Dem want-Impfv.F be.Prs.Sg that I Dubai go-Sbjv.1Sg\\
    \glt `She wants that I go to Dubai'
    \end{xlist}
\end{exe}

Now, however, only the desire meaning of `ca:h' 
remains and it takes an infinitival/subjunctive 
complement.

\subsection{Loss of a particular aspectual form}
Consider the {\em -(y)a: kar} construction in
contemporary Hindi-Urdu. This construction combines with a range of aspects.

\begin{exe}
    \ex 
    \begin{xlist}
        \ex Imperfective:
         \gll vo ro:z sku:l ja:-ya: kar-ti: hai\\
              Dem  everyday school go.Pfv.M habit be.Impfv.F be.Prs.Sg\\
            \glt `She goes to school every day'   
        \ex Subjunctive:  
        \gll mEN ca:h-ta: hu:~ ki vo ro:z sku:l ja:ya: kare\\
            1Sg want-Impfv.M be.Prs.1Sg that Dem everyday school go.Pfv.M do.Impfv.3Sg.Subjv\\
            \glt `I want that she/he go to school everyday'
        \ex Future: 
        \gll agle mahi:ne se vo ro:z sku:l ja:-ya: kare gi:\\
    next month from Dem everyday school go.Pfv.M do.Impfv.3Sg.Subjv Foc.Fut\\
\glt `From next month, she will go to school everyday'
        \ex Infinitive: 
        \gll us-e ro:z sku:l ja:-na: qat'ii pasand nahĩ hai\\
    3Sg-Dat everyday school go-Inf absolutely like not be.Prs.Sg\\
\glt `She absolutely does not like going to school everyday'

        \ex  Polite Imperative:
        \gll meher-ba:-ni: kar ke kal sku:l ja:-iye ga:\\
    kindness do having tomorrow school go-Imp.Pol Pol.Part\\
        \glt `Please do go to school tomorrow'
        \ex Plain Imperative:
        \gll waqt se sku:l ja:ya: kar\\
            time from school go.Pfv.M do.Impfv.2Sg.Imp\\
        \glt `Go to school on time'
        \ex Future Imperative:
        \gll ro:z sku:l ja:-ya: kar-na:\\
            daily school go.Pfv.M do-Inf\\
        \glt `To go to school daily'
    \end{xlist}
\end{exe}

However, it does not combine with the perfective and progressive.
\begin{exe}
    
    \ex \label{ex:jayaki}\ *Perfective:
    \gll vo ro:z sku:l ja:ya: ki:\\
    Dem everyday school go.Pfv.M did.Pfv.F.Sg\\
\glt `She used to go to school everyday.'
\end{exe}

\begin{exe}
    \ex \ *Progressive:
    \gll vo ro:z sku:l ja:ya: rah-i: hai\\
    Dem everyday school go.Pfv.M stay-Impfv.F be.Prs.Sg\\
\glt `She goes to school everyday (as a routine).'
\end{exe}

The ungrammaticality of the perfective form is a recent development in the history of the language. \citet{mcgregor1972} excellent Hindi grammar, published in 1972, reports instances of the -\textit{ya: kar} construction in the perfective but notes that they are rare (p. 137). In the 50 years since, the combination has gone from rare to non-existent for contemporary speakers, while the construction remains alive and well in other tenses. A further reason for not entertaining a semantic incompatibility argument for the ungrammaticality of the perfective is that the semantically close -\textit{ta: rah-} `-Impfv stay-' construction freely combines with the perfective, as shown in Section~\ref{sec:taraha}.

The impossibility with the progressive can perhaps
be explained on semantic grounds. However, the
impossibility of the combination with the perfective remains to be explained and is the subject of this talk.

To understand the current constraints on the -ya: kar construction, particularly its incompatibility with the perfective aspect, we must consider its historical evolution.  The examples(with book title, author, and year of publication indicated in parentheses) presented below, drawn from this period, offer concrete instances of the perfective aspect occurring with the -ya: kar construction, directly contradicting its modern-day impossibility.


\begin{exe}
    \ex \label{ex:old_jayaki} ja:-ya: k\~{i}: (Khurshid Bahu, Mirza Hadi Ruswa)
    \gll tum  aur  pya:-ri:  sa:-ji-dah  is  maka:n  meN  a:-ya:  ja:-ya:  k\~{i}:\\
    2Sg.Pron.Inf  and  dear.F.Sg  Sajida  this  house  in  come-Pfv.M  go-Pfv.M  did.Impfv.Pl\\
\glt `You and dear Sajida used to come and go in this house.'
\ex rakh-a: k\~{i} (Umrao Jan Ada, Mirza Hadi Ruswa, 1899)
\gll mere  na:m=ka:  ta'ziyah  kha:-num  mar-te  dam  tak  rakh-a:  k\~{i}\\
    my  name=GEN  taziyah  Khanum  die-Impfv.M  breath  until  keep-Pfv.M.Sg  do-Impfv.Pl\\
\glt `Khanum kept the taziyah of my name until her last breath.'

    \ex  taR-pa: ki: (Fasana-e-Ajaib,Rajab Ali Baig Suroor,1844)
    \gll yeh sun kar tama:m shab taR-pa: ki:\\
    this hear having all night writhe-Pfv.M.Sg did.Pfv.F.Sg\\
     \glt `Having heard this, she writhed all night.'
     
       \ex  de-kha:  kiye (Urdu-e-Mualla, Mirza Ghalib, 1869)
    \gll agast=ke  mahi:ne=ka:  ha:l  kuchh  ma'lu:m  nahĩ  kal  sha:m=ko  do  do  moonDhe  rakh  kar  ka-i:  a:d-mi:  de-kha:  kiye\\
    August=GEN  month=GEN   state  some  known  not  yesterday  evening=GEN  two  two  stool  keep  having  several  person  see.Pfv.M  did.Impfv.Pl\\
    \glt `The situation of the month of August is unknown-just last evening, several people were seen stacking two stools on top of each other.'
    
    \ex soca: ki: (Ayama, Deputy Nazir Ahmad, 1891)
    \gll akel-i: paR-i: kuchh socha: ki:\\
    alone-F fall-Pfv. some think.Pfv.M did.Pfv.F\\
\glt `Lying alone, she was thinking something.'

\end{exe}

The contrast between the modern ungrammaticality in \ref{ex:jayaki} and the historical attestations in \ref{ex:old_jayaki}  establishes a clear case of diachronic change resulting in a paradigm gap. This shift requires an explanation. The following sections will argue that this gap arose from a fundamental morphosyntactic conflict within the construction itself, specifically related to case assignment and the invariant nature of the \textit{-ya:} participle.

\section{The {\em -ya: kar} construction}

\subsection{Semantics}

The historical usage of the perfective \textit{-ya: kar} construction reveals a distinct and specific semantic profile, characterized by two core features: (i) its strong association with temporal adverbials that define a bounded timeframe,in many—though not all—cases, the construction co-occurs with durational adverbs like `ra:t bhar', `der tak', and (ii) its strong preference for atelic predicates. The convergence of these features created a specific aspectual niche that we term \textit{bounded atelicity} \citep{atelic}.

As illustrated by the examples presented below (with book titles, authors, and years of publication indicated in parentheses), this construction frequently appears alongside durational adverbials in literary texts.

\begin{exe}
 \ex ra:t bhar \textit{(Umrao Jan Ada, Mirza Hadi Ruswa, 1899)}
    \gll meN khud ra:t bhar ro-ya: ki:\\
    1Sg self night full cry-Pfv.M.Sg did.Pfv.F.Sg\\
\glt `I myself cried all night.'

\ex der tak\textit{( Ahmaqullazin, Munshi Sajjad Hussain, 1906)}
\gll der tak dukh dard ka-ha: kiye\\
    long until sorrow pain say.Pfv.M.Pl did.Impfv.Pl\\
\glt `They kept talking about sorrow and pain for a long time.'

\ex ghaNton~ \textit{(Firdaus-e-Barin,  Abdul Halim Sharar, 1899)}
\gll aur khud bhi: te:re: sa:th ghaNton~ khaR-ii: ro-ya: ki:\\
    and self also.Part your.2Sg.Poss with.Postp hours stand-Pfv.F cry-Pfv.M.Sg did.Pfv.F.Sg\\
\glt `And I myself also stood and cried with you for hours.

\end{exe}

A list of adverbs used in the construction is provided below.
{\small
\begin{multicols}{3}
\begin{enumerate}
  \item chay sa:l tak
  \item der tak
  \item is samay
  \item Ek mint
  \item bar-s\~{u}
  \item badi: de:r
  \item bohat de:r tak
  \item pa:nc pehar
  \item tama:m shab
  \item ca:r din
  \item cand lamh\~{o}
  \item cand mint
  \item din bhar
  \item din caRhe
  \item do sa:l tak
  \item der tak
  \item ra:t bhar
  \item ra:t ko
  \item sa:re: sa:re: din
  \item shab ko
  \item sadiy\~{o} talak
  \item kuch din\~{o}
  \item kuch der
  \item kal sha:m
  \item guzray huay bars
  \item ghaRiy\~{o}
  \item ghant\~{o}
  \item marte: dam tak
  \item mahi:-n\~{o}
\end{enumerate}
\end{multicols}
}
This specific semantic niche—the quantification of an atelic activity within a bounded timeframe—is key to understanding the construction's eventual loss. Its functional space overlapped with two more general and stable constructions:
1.  The simple imperfective for habituality (e.g., \textit{ja:-ti thi} `used to go').
2.  The simple perfective for bounded atelic events, often with a durational adverb (e.g., \textit{vo ra:t bhar roi} `she cried all night').

We posit that this very specific semantic role made the perfective \textit{-ya: kar} form functionally redundant. When the morphosyntactic conflict (Section~\ref{sec:syntax}) placed pressure on the construction, its narrowly defined meaning provided no functional incentive for speakers to preserve it. The language already had simpler, unproblematic ways to express the same concept of a bounded atelic event, leading to the construction's disappearance from the perfective paradigm.

\subsection{Syntax}\label{sec:syntax}

The historical data reveals two critical and consistent syntactic properties of the perfective \textit{-ya: kar} construction that are fundamental to explaining its eventual loss. These properties directly conflict with the core rules of Urdu grammar governing perfective transitives, creating an unstable configuration for speakers.

\textbf{1. Nominative Case Assignment.}
Despite containing a transitive verb and a perfective auxiliary, the construction consistently requires its subject to be in the \textbf{nominative case}. This violates the standard rule of Urdu grammar where a transitive verb in the perfective aspect assigns \textbf{ergative} case to its subject \citep{ErgativityMiriam}. This holds true even when the embedded verb is clearly transitive, as shown in examples (\ref{ex:kiya_ki}) and (\ref{ex:rakha_ki}) below.

\textbf{2. Invariant Participle.}
The \textit{-ya:} element, a perfective participle, remains morphologically \textbf{invariant}. It is frozen in the default masculine singular form and does not agree in gender, number, or case with the subject or the object. This contrasts sharply with simple perfective verbs, which must agree with the object in gender and number.

The following examples from literary texts illustrate these properties. Critically, the subjects (e.g., \textit{ja'fari:}, \textit{Xa:-nam}) are nominative, and the participles (\textit{ka-ha:}, \textit{rakh-a:}) are invariant masculine singular forms, even when the context or the auxiliary's agreement suggests a feminine or plural subject.

\begin{exe}
\ex\label{ex:kiya_ki} ki:-ya: ki: (Fasana-e-Ajaib, Rajab Ali Baig Suroor, 1844)
\gll der tak akhlaaq o muhabbat malika:=ka: mazku:r kiya: ki:\\
    long until ethics and love Queen=GEN mention do.Pfv.M.Sg did.Pfv.F.Sg\\
\glt `For a long time, she kept mentioning Queen's ethics and love.'

\ex jhap-ka:-ya: kiye (Aag ka Deriya, Qurat-ul-Ain Haider,1959)

    \gll dono kuch der tak cup ca:p an-dhere men pal-ken jhap-kaya: kiye \\
    both a while silently darkness eyelids blink.Pfv.M did.Imfv.pl  \\
    \glt 'Both blinked their eyelids for a while in the darkness in silence.'

\ex\label{ex:kaha_ki}    ka-ha:  ki: (Akhteri Begum, Mirza Hadi Ruswa)
\gll ja'fari:  ap-ni:  Dhi-Ta:-i:  se  jhu:T  ka-ha:  ki:\\
    Jafari  self-Poss.F  stubbornness-F  from  lie  say-Pfv.M.Sg  did-Pfv.F.Sg\\
\glt `Jafari, with her stubbornness, kept telling lies.'
\ex rakh-a: ki-en (Umrao Jan Ada, Mirza Hadi Ruswa, 1899)
\gll\label{ex:rakha_ki} mere  na:m=ka:  ta'ziyah  kha:-num  mar-te  dam  tak  rakh-a:  ki:-en:\\
    my  name=GEN  taziyah  Khanum  die-Impfv.M  breath  until  keep-Pfv.M.Sg  do-Impfv.Pl\\
\glt `Khanum kept the taziyah of my name until her last breath.'
\ex  de-kha:  kiye (Urdu-e-Mualla, Mirza Ghalib, 1869)
    \gll agast=ke  mahi:ne=ka:  ha:l  kuchh  ma'lu:m  nahĩ  kal  sha:m=ko  do  do  moonDhe  rakh  kar  ka-i:  a:d-mi:  de-kha:  kiye\\
    August=GEN  month=GEN   state  some  known  not  yesterday  evening=GEN  two  two  stool  keep  having  several  person  see.Pfv.M  did.Impfv.Pl\\
    \glt `The situation of the month of August is unknown-just last evening, several people were seen stacking two stools on top of each other.'

\ex  de-kha:  kiye (Urdu-e-Mualla, Mirza Ghalib, 1869)
    \gll agast=ke  mahi:ne=ka:  ha:l  kuchh  ma'lu:m  nahĩ  kal  sha:m=ko  do  do  moonDhe  rakh  kar  ka-i:  a:d-mi:  de-kha:  kiye\\
    August=GEN  month=GEN   state  some  known  not  yesterday  evening=GEN  two  two  stool  keep  having  several  person  see.Pfv.M  did.Impfv.Pl\\
    \glt `The situation of the month of August is unknown-just last evening, several people were seen stacking two stools on top of each other.'
\ex so-ca: ki: (Ayama, Deputy Nazir Ahmad, 1891)
    \gll akel-i: paR-i: kuch so-ca: ki:\\
    alone-F fall-Pfv. some think.Pfv.M did.Pfv.F\\
\glt `Lying alone, she was thinking something.'

\ex pi:-sa: kiya: (Safeena-e Gham-e Dil, Qurat-ul-Ain Haider, 1953)
\gll do  saal  tak  mEN  cakki:  pi:-sa:  ki-ya:  aur  meN-ne  navaaRen  bu-nĩ\\
    two  year  until  1Sg  grinding.stone  grind.Pfv.M.Sg  did.Impfv.Sg  and  1Sg-Erg  tapes  weave-Pfv.F.Pl\\
\glt `For two years, I ground the millstone and I wove the yarn.'
\end{exe}

These two properties—obligatory nominative subjects and an invariant participle—define the syntactic signature of the construction. However, they also created a fundamental conflict for speakers. The language's core grammar simultaneously demanded that a perfective transitive clause have an ergative subject. This placed the construction under immense syntactic pressure. We propose that this underlying instability is the primary cause for its disappearance from the paradigm; speakers resolved the conflict by avoiding the construction entirely in the perfective, where the case assignment paradox was most acute.




\section{The {\em -ta: rah-} construction: A robust alternative}\label{sec:taraha}

A crucial question arising from the paradigm gap in \textit{-ya: kar} is whether it led to a loss of expressive power in the language. This can be examined by analyzing the \textit{-ta: rah-} (`keep V-ing') construction, which exhibits no such gap and serves as a functional alternative for expressing sustained or continuous action. Its robustness across the entire aspectual paradigm is demonstrated below using the verb \textit{ja:-} ('to go').

\begin{exe}
    \ex 
    \begin{xlist}
    \ex Perfective: 
        \gll vo cup ca:p sku:l ja:-ti: rah-i:\\
        Dem silent silently school go-Impfv.F stay-Impfv.F\\
        \glt `She kept going to school silently.'

    \ex Imperfective: 
        \gll vo sku:l ja:-ti: rah-ti: hai\\
        Dem school go-Impfv.F stay-Impfv.F be.Prs.Sg\\
        \glt `She keeps going to school .'
    
        \ex Subjunctive: 
            \gll meN ca:h-ta: hu:n ki vo ek sa:l tak sku:l ja:-ti: rah-e\\
    1Sg want-Impfv.M be.Prs.1Sg that Dem one year until school  go-Impfv.F stay-Impfv.F\\
\glt `I want that she keeps going to school for one year.'
        
        \ex Future:  

        \gll vo das sa:l  tak sku:l ja:-ti: rah-e gi:\\
            Dem ten years until school go-Impfv.F stay-Impfv.F Fut.3Sg.F\\
        \glt `She will keep going to school for ten years.'

        \ex Infinitive:  
       \gll sku:l ja:-te reh-na: us=ki: a:dat hai\\
     school go-Inf 3Sg-Poss habit be.Prs.Sg\\
\glt `Going to school is her habit.'

        \ex  Polite Imperative:
        \gll ro:za:na:  sku:l  ja:-ti:   rah-iye  ga:\\
    daily  school go-Impfv.F  stay-Impv.Pol  Pol.Part\\
\glt `Please keep going to school daily.'
        \ex Plain Imperative 
        \gll cup  kar  aur  sku:l ja:-ti:  rah\\
    silent  do.Imp.Inf  and school  go-Impfv.F  stay.Imp.Inf\\
\glt `Be quiet and keep going to school.'
        \ex Future Imperative 
        \gll tum  ek  sa:l sku:l  ja:-ti:  rah-na:\\
    2Sg  one  year school  go-Impfv.F  stay-Inf\\
\glt `You keep going to school for one year.'

    \end{xlist}
\end{exe}

The stark contrast between the two constructions is revealed in a minimal pair. The meaning intended by the lost perfective \textit{-ya: kar} form is readily expressed by the perfective form of \textit{-ta: rah-}.

\begin{exe}
    \ex
    \begin{xlist}
        \ex Lost example
        \gll vo ro:z sku:l ja:-ya: ki:\\
Dem everyday school go.Pfv.M did.Pfv.F.Sg\\
\glt She used to go to school everyday.' 

\ex Counterpart with '-ta: rah-'
\gll vo ro:z sku:l ja:-ti: rah-i:\\ 
Dem everyday school go-Impfv.F stay-Pfv.F.Sg\\ 
\glt She used to keep going to school everyday.'
       
    \end{xlist}
\end{exe}



The "-ta: rah-" construction, unlike "-ya: kar", readily combines with the perfective aspect to express a continuous or sustained action in the past, similar to the intended meaning of the lost -ya: kar-Pfv form. This suggests that the semantic function associated with a past habitual or continuous action is still expressible in the language through an alternative grammatical construction.

\section{Why the gap?}
\cite{ileri2018} propose that the observed paradigm gap within Turkish desiderative constructions in the third-person plural arises from a confluence of factors. Firstly, the near-absence of attested instances of the -AsI + 3PL suffix combination in corpus data suggests a paucity of positive evidence available to learners. This lack of direct input, while not explicitly negative, may effectively function as implicit negative evidence, hindering the acquisition and production of the expected forms. Secondly, the presence of alternative, competing forms for the expression of third-person plural desideratives further contributes to the marginalization and eventual paradigm gap of the -AsI construction. The availability of these alternative forms reduces the necessity and, consequently, the frequency of the targeted construction, ultimately leading to its diminished usage and potential grammaticalization of the gap. Following the approach, we argue that the loss of the perfective \textit{-ya: kar} form in Urdu can be attributed to a fundamental \textbf{syntactic conflict}. This conflict made the construction unstable, and its functional replaceability (shown in Section~\ref{sec:taraha}) allowed it to be abandoned without loss of expressivity.

\subsection{Motivating the gap in Urdu}

The diachronic absence of the perfective \textit{-ya: kar} construction is best explained by a morphosyntactic paradox. The construction's internal requirements directly contradict a core rule of Urdu grammar.

\begin{enumerate}
    \item \textbf{General Rule:} A transitive verb in the perfective aspect (\textit{kar}) requires its subject to be in the \textbf{ergative} case.
    \item \textbf{Construction-Specific Rule:} The \textit{-ya: kar} construction, even in its perfective historical form, consistently requires its subject to be in the \textbf{nominative} case, as evidenced by the examples presented earlier (e.g., the subject \textit{ja'fari:} in (\ref{ex:kaha_ki}), or \textit{Xa:-nam} in (\ref{ex:rakha_ki}))s.
\end{enumerate}

This conflict is not merely theoretical. It is exacerbated when the embedded verb (\textit{-ya:}) is itself transitive. In such cases, \textbf{both verbs semantically imply an agentive subject}, yet the construction forbids the ergative case that would typically mark that agent.  For instance, in the Example \ref{ex:kiya_ki} the embedded verb "kiya:(mentioned)" is transitive, yet its implied subject(the narrator) remains nominative with the "-ya:kar" frame. This creates a unresolvable tension for the speaker.
This clash in case assignment expectations for a single underlying subject likely rendered the perfective "-ya: kar" construction grammatically unstable, leading to its avoidance by speakers. The availability of alternative means to express habitual past actions (as seen in the modern avoidance and potential use of simple perfective with adverbs) further facilitated this disuse, resulting in the observed paradigm gap in contemporary Hindi-Urdu.

Our analysis of historical attestations provides quantitative support for this point of tension. The construction appeared with both transitive and intransitive verbs as shown in Table \ref{tab:frequency_table_books_intransitive}, Table \ref{tab:frequency_table_books_transitive} and Table \ref{tab:frequency_table_books_both}.

\begin{longtable}{llclp{4cm}} 
\caption{Frequency and Transitivity of Verbs with "Kar" Forms}
\label{tab:frequency_table_books_intransitive} \\ 
\hline
\textbf{Verb} & \textbf{"Kar" Form} & \textbf{Count} & \textbf{Transitivity} & \textbf{Counter Example} \\ \hline
\endhead 
\hline
\multicolumn{5}{r}{{Continued on next page}} \\ 
\endfoot 
a:-ya: & ki: & 1 & intransitive & \\ \hline
ik-tafa: & ki: & 1 & intransitive & \\ \hline
ba-ja: & kiye & 1 & intransitive & \\ \hline
bad-la: & kiya: & 1 & intransitive & \\ \hline

bar-sa: & ki: & 1 & intransitive & \\ \hline
bha-ra: & kiye & 1 & intransitive &  it can be transitive e.g mEn ne gla:s men pani: bha-ra: \\ \hline
bhun-bha-naya: & ki: & 1 & intransitive & \\ \hline
ba-ha: & kiya: & 1 & intransitive & it can be transitive e.g taiz hawa dhu:l ba-ha: le gayi:\\ \hline
phi-ra: & ki: & 1 & intransitive & \\ \hline
phi-ra: & kiya: & 1 & intransitive & \\ \hline
taR-pa: & ki: & 1 & intransitive & \\ \hline
taR-pa: & kiya: & 2 & intransitive & \\ \hline
te-ra: & kiya: & 3 & intransitive & \\ \hline
te-ra: & kiye & 1 & intransitive & \\ \hline
Tak-ra:-ya: & ki: & 1 & intransitive & \\ \hline
Tam-Tama-ya: & k\~{i} & 1 & intransitive & \\ \hline
Teh-la: & ki: & 1 & intransitive & \\ \hline
Teh-la: & kiye & 1 & intransitive & \\ \hline
ja-ya: & k\~{i} & 1 & intransitive & \\ \hline
ja-ga: & ki: & 1 & intransitive & \\ \hline
jhil-mi:-laya: & ki: & 1 & intransitive & \\ \hline
car-ca-raya: & k\~{i} & 1 & intransitive & \\ \hline
ca-la: & ki: & 1 & intransitive & \\ \hline
ca-la: & kiya: & 1 & intransitive & \\ \hline
ca-la: & k\~{i} & 1 & intransitive & \\ \hline
ca-la: & kiye & 1 & intransitive & \\ \hline
ro-ya: & ki: & 7 & intransitive & it can be transitive e.g. wo ap-ni: qismat ko roya:\\ \hline
ro-ya: & kiya: & 1 & intransitive & \\ \hline
ro-ya: & k\~{i} & 1 & intransitive & \\ \hline
so-ca: & k\~{i} & 1 & intransitive & \\ \hline
so-ya: & ki: & 1 & intransitive & \\ \hline
so-ya: & kiya: & 1 & intransitive & \\ \hline
kha:-ya: & ki: & 1 & intransitive & \\ \hline
khe-la: & ki: & 1 & intransitive & it can be transitive e.g ham ney krikaT ka mac khe-la:\\ \hline
ki-ya: & kiye & 3 & intransitive & \\ \hline
ga-ya: & ki: & 1 & intransitive & \\ \hline
ga-ya: & kiye & 1 & intransitive & \\ \hline
guz-ra: & kiye & 1 & intransitive & \\ \hline
gh\~{u}-ja: & k\~{i} & 1 & intransitive & \\ \hline
la-ya: & kiye & 1 & intransitive & it can be transitive e.g. wo seb la-ya:\\ \hline
mus-kara:-ya: & k\~{i} & 1 & intransitive & \\ \hline
na:-ca: & ki: & 1 & intransitive & \\ \hline
h\~{a}-sa: & ki: & 1 & intransitive & \\ \hline
hu-wa: & kiya: & 1 & intransitive & \\ \hline
hu:-wa: & k\~{i} & 4 & intransitive & \\ \hline

\end{longtable}

\begin{longtable}{llclp{4cm}}

\caption{Frequency and Transitivity of Verbs with "Kar" Forms}
\label{tab:frequency_table_books_transitive} \\ 
\hline
\textbf{Verb} & \textbf{"Kar" Form} & \textbf{Count} & \textbf{Transitivity} & \textbf{Counter Example} \\ \hline
\endhead 
\hline
\multicolumn{5}{r}{{Continued on next page}} \\ 
\endfoot 

baja:-ya: & kiya: & 1 & transitive & \\ \hline

ba-ha-ya: & ki: & 1 & transitive & \\ \hline
beh-la:-ya: & ki: & 1 & transitive & \\ \hline
pa-Rha: & ki: & 1 & transitive & \\ \hline
ph\~{u}-ka: & kiye & 1 & transitive & \\ \hline
phe-ra: & k\~{i} & 1 & transitive & \\ \hline
peh-na: & k\~{i} & 1 & transitive & \\ \hline
pe-Ta: & k\~{i} & 1 & transitive & \\ \hline
pe-sa: & kiya: & 1 & transitive & \\ \hline
ta-ka: & kiya: & 1 & transitive & it can be intransitive e.g. ba-cey ney bahir taka:\\ \hline
jhap-ka-ya: & kiye & 1 & transitive & \\ \hline
che-Ra: & kiya: & 1 & transitive & \\ \hline
de-kha: & ki: & 7 & transitive & it can be intransitive e.g mEN ney idhar udhar de-kha \\ \hline
de-kha: & kiya: & 8 & transitive & \\ \hline
de-kha: & kiye & 8 & transitive & \\ \hline
ra-kha: & k\~{i} & 1 & transitive & \\ \hline
so-na: & ki: & 3 & transitive & \\ \hline
so-na: & kiya: & 4 & transitive & \\ \hline
so-na: & k\~{i} & 3 & transitive & \\ \hline
so-na: & kiye & 1 & transitive & \\ \hline
so-na:-ya: & kiya: & 1 & transitive & \\ \hline
so-ca: & ki: & 2 & transitive & \\ \hline
ka-ha: & ki: & 1 & transitive & it can be intransitive e.g us ney bas y\~{u}hi: ka-ha:\\ \hline
ka-ha: & k\~{i} & 1 & transitive & \\ \hline
ka-ha: & kiye & 2 & transitive & \\ \hline
ki-ya: & ki: & 1 & transitive & \\ \hline
ghu-ra: & kiye & 1 & transitive & \\ \hline
la-Ra: & k\~{i} & 1 & transitive &  it can be intransitive e.g wo ba-ha:dri: sey la-Ra:\\ \hline
la-Ra: & kiye & 1 & transitive & \\ \hline
la-ga:-ya: & ki: & 1 & transitive & \\ \hline
la-ga:-ya: & kiye & 1 & transitive & \\ \hline
ma:-ra: & kiye & 1 & transitive & \\ \hline
mi-la: & kiye & 1 & transitive & it can be intransitive e.g wo mujh sey mi-la:  \\ \hline

\end{longtable}

\begin{table}[H]
\centering
\caption{Verbs appeared both transitive and intransitive }
\label{tab:frequency_table_books_both} 
\begin{tabular}{llc}
\hline
\textbf{Verb} & \textbf{"Kar" Form} & \textbf{Transitive Count / Intransitive Count}  \\ \hline

so-ca: & kiye & 2 /2  \\ \hline
so-ca: & kiya: & 1 /1  \\ \hline
kha:-ya: & kiye & 1/1   \\ \hline

\end{tabular}
\end{table}

Furthermore, two verbs appeared in both transitive and intransitive frames as shown in Table \ref{tab:frequency_table_books_both} (e.g., \textit{so-ca:} `thought (about something)' vs. `thought'), demonstrating that the conflict was not limited to a few lexical items but was a pervasive structural feature of the construction.

\subsubsection{Dil kar}
Here too we have "kar" which does not license an ergative subject

\begin{exe}
    \ex 
     \begin{xlist}
    \ex
    \gll mera:  dil  kiya:  ke  mEN  abhi:  cala:  ja:-\~{u}:~\\
    1Sg.Poss  heart  did  that  1Sg  now  move  go-Subjv.1Sg\\
\glt `My heart desired that I leave right now.'
     \ex
    \gll us  waqt  mera:  ba:har  ja:ne=ko  bahut  dil  kiya:\\
    that  time  1Sg.Poss  outside  go.Inf=GEN  very  heart  do.Pfv.M.Sg\\
\glt `At that time, I really wanted to go out.'
\ex
\gll sardi:  itni:  zi-ya:-da:  thi:  ke  mera:  garm  ca:e  pi:-ne=ko  dil  kiya:\\
    cold  so.much  much  was  that  1Sg.Poss  hot  tea  drink-Inf  to  heart  do-Pfv.M.Sg\\
\glt `The cold was so much that my heart wanted to drink hot tea.'

\ex
\gll us  ka:  dil  kiya:  ke  vo  zor  se  h\~{a}se\\
    3Sg.Poss  heart  did  that  3Sg  force  with  laugh-Subjv.3Sg\\
\glt `His/Her heart desired that he/she laugh loudly.'

    \end{xlist}
\end{exe}

In all above examples ending with "dil kiya:", the grammatical subject of the verb  "kiya:" is "dil"(heart). This subject is not marked with the ergative postposition  "ne".
 This consistent absence of ergative marking leads to the following thoughts.


\begin{enumerate}
   \item The "dil kiya:" construction lacks a prototypical external argument (agent). As ergative marking in Urdu is typically licensed by agentive subjects of transitive verbs in the perfective aspect, the absence of such an argument in these constructions means the conditions for ergative assignment do not arise.
   \item Perhaps it is relevant that "dil kiya:" construction, functioning as a more lexicalized Noun+Verb compound to express desire, exhibits less syntactic articulation compared to the "-ya:kar" structure, which is a more transparent syntactic combination of a perfective verb stem and an auxiliary to mark habitual aspect. This difference in structural composition and grammatical function likely contributes to why "kar" in "dil kiya:" does not license an ergative subject.
\end{enumerate}

\subsubsection{An implication for learning}

Our diachronic account raises a significant synchronic question: how is this paradigm gap sustained in the mental grammar of contemporary Hindi-Urdu speakers?

For simple verbs, speakers can productively apply aspectual morphology to novel or nonce words (e.g., they know that the perfective of a hypothetical verb \textit{zimb} would be \textit{zimb-aa}), suggesting that paradigm gaps for simplex verbs are rare and perhaps learned by negative evidence.

However, the \textit{-ya: kar} construction is a complex, periphrastic pattern. We propose that the combinations of such complex constructions with specific aspects are likely learned not by productive rule but from \textbf{positive evidence} in the input. The modern gap persists because learners are exposed to no positive evidence for the perfective form. The absence of \textit{*-ya: ki:} in the input is interpreted as a systematic gap, not as an accident of frequency. This is bolstered by the presence of fully grammatical alternatives (\textit{-ta: rah-}, simple perfective), which provide no functional motivation for learners to hypothesize the missing form.

Thus, the gap is sustained because the necessary triggering evidence for this specific combination is absent from the primary linguistic data, and the grammar has stabilized in a state where the construction is simply undefined for the perfective aspect.

\subsection{-Ya: Kar-PFv frequency: 19th century books vs. web samples}

To compare the prevalence of the \textit{-Ya:Kar}-Perfective (PFv) pattern in 19th-century Urdu books and contemporary Urdu web samples, we analyzed two distinct datasets: a sample of 133 sentences (Sample-A) and contemporary data is drawn from the UMass Urdu Corpus  \cite{umassurducorpus}, a collection of approximately 735 million tokens, encompassing diverse web-based text from sources such as Common Crawl, Twitter, and news articles.

The initial analysis of Sample-A identified 80 unique surface realizations of the \textit{-Ya: Kar}-PFv construction (e.g., \textit{beh-la:-ya: ki:}, \textit{soya: ki:}). A subsequent search for these specific patterns in Umass Urdu corpus yielded no exact matches, necessitating a shift in our approach to the contemporary data.

We then randomly selected 100 sentences from the UMass Urdu Corpus (Sample-B) that contained verb forms fulfilling a similar semantic role – namely, Perfective Aspect (Pfv) combined with auxiliaries resembling \textit{karna:} ( \textit{kiyā} / \textit{ki:} - to do/make) or its conjugated forms. These included \textit{karti:} ( \textit{karti:} - does/makes [fem. sg.]), ( \textit{karte} - do/make [masc. pl./respectful sg.]), and the bare perfective form (e.g., \textit{soya:}  - slept). 
For instance, the structure in (\textit{ra:j kapu:r aksar kaha: karte thay ki dili:p kuma:r ki: sha:di: par woh ghuṭnon ke bal un ke ghar jā\~e ge}) from contemporary usage conveys a past habitual action similar to the function observed in older \textit{-Ya: Kar} forms (which might have been expressed as "\textit{ra:j kapu:r aksar kaha: kiye ki dili:p kuma:r ki: sha:di: par woh ghuṭnon ke bal un ke ghar jā\~e ge}").  The rationale was that even if the specific \textit{-Ya: Kar} combination was no longer prevalent, contemporary Urdu likely employs other strategies to express completed or habitual past actions involving a \textit{kar}-related auxiliary. 
Each of these 100 sentences was manually coded for verb transitivity and subject overtness to enable a comparison of grammatical feature distribution between the two datasets, despite the evolution in the specific surface forms conveying this semantic function.

The raw frequency counts for the categorized verb patterns in Sample-A and Sample-B are presented in Table~\ref{tab:raw_frequency_yaa_kar_pfv}.

\begin{table}
    \centering
    \caption{Raw Frequency Count of \textit{-Ya: Kar}-PFv in 19th Century Books (Sample-A) and Verb+\textit{kar} Forms in Web Samples (Sample-B)}
    \label{tab:raw_frequency_yaa_kar_pfv}
    \begin{tabular}{lccc}
        \hline
        \textbf{Transitivity} &\textbf{Subject} & \textbf{Sample-A} & \textbf{Sample-B} \\
        \hline
        Intransitive&overt & 51 & 31 \\
        Intransitive&covert  & 15 & 11 \\
        Transitive& overt & 41& 45 \\
        Transitive& covert & 25 & 11 \\
        \hline
    \end{tabular}
\end{table}
To account for the different sizes of the two samples, relative frequencies were calculated to allow for a direct and proportional comparison of the pattern distribution across the corpora. The relative frequencies for each category are shown in Table~\ref{tab:relative_frequency_yaa_kar_pfv}.

\begin{table}
    \centering
    \caption{Relative Frequency of \textit{-Ya: Kar}-PFv in 19th Century Books (Sample-A) and Verb+\textit{kar} Forms in Web Samples (Sample-B)}
    \label{tab:relative_frequency_yaa_kar_pfv}
    \begin{tabular}{lccc}
        \hline
        \textbf{Transitivity} &\textbf{Subject} & \textbf{Sample-A} & \textbf{Sample-B} \\
        \hline
        Intransitive&overt & 38.6\% & 33.0\% \\
        Intransitive&covert  & 11.4\% & 11.0\% \\
        Transitive& overt & 31.1\% & 45.0\% \\
        Transitive& covert & 18.9\% & 11.0\% \\
        \hline
    \end{tabular}
\end{table}

Table~\ref{tab:relative_frequency_yaa_kar_pfv} compares the distribution of verb patterns based on transitivity and subject overtness in 19th-century Urdu books and contemporary web samples. The higher relative frequency of transitive verbs with overt subjects in the web sample (45.0\%) compared to the book sample (31.1\%) suggests a potential shift in the preference for explicit agents in transitive constructions between these two periods. 

Further investigation into the specific auxiliary verbs and their semantic roles in both datasets could provide more detailed insights into these linguistic differences between older and contemporary Urdu.
\section{Human evaluation}
\subsection{Experiment}

This study was conducted in accordance with the principles outlined in the Declaration of Helsinki. The primary aim of this evaluation was to obtain direct, quantitative evidence from native speakers on the acceptability of the perfective \textit{-ya: kar} construction in modern Urdu. Our hypothesis was that sentences containing the perfective form (e.g., \textit{ro-ya: ki:}) would be judged as sharply unnatural compared to their modern counterparts expressing habitual past action.

\subsubsection{Participants}
Twenty-one native speakers of Urdu (11 female, 10 male) were recruited for the study. Participant age ranged from 20 to 48 years . All participants reported having completed higher education. Participation was voluntary.

\subsubsection{Design and procedure}
A forced-choice acceptability judgment task was designed. The stimulus set consisted of 20 unique sentences historically attested in 19th-century literature, featuring the perfective \textit{-ya: kar} construction. For each historical sentence, a modern counterpart was created by replacing the perfective \textit{-ya: kar} form with a grammatical alternative, most commonly the imperfective habitual form (\textit{-ta: tha:/ti: thi:}).

Participants were presented with these 20 minimal pairs in random order. For each pair, they were asked the question: “Which sentence sounds more natural to you?” . They were forced to choose one of the two options. The experiment was administered digitally.

\subsubsection{Results}
The results were unequivocal and aligned perfectly with our hypothesis. Across all 20 items and 21 participants, generating 420 total data points, \textbf{not a single participant} selected the sentence containing the perfective \textit{-ya: kar} construction as the more natural option. This represents a 0\% acceptance rate for the historical form. All 420 responses (100\%) favored the modern grammatical alternative.

This stark, categorical rejection of the perfective form provides powerful convergent evidence that supports our corpus findings and confirms that a robust paradigm gap exists in the modern grammar of Urdu speakers. The complete absence of any selection for the target construction indicates that its ungrammaticality is not a subtle or gradient effect but a definitive and stable feature of the contemporary linguistic system.

\section{Conclusion}
This study has demonstrated that the contemporary absence of the perfective -yā: kar construction in Hindi-Urdu is a  paradigm gap resulting from grammatical change. Historical evidence from the late 19th century attests to its productive use, a usage characterized by its unique morphosyntactic signature: a nominative subject despite transitive embedded verbs and an invariant -yā: participle. We have argued that this very configuration created a fundamental syntactic conflict with the core ergative alignment of the perfective aspect in Hindi-Urdu. The convergence of evidence from historical texts, large-scale corpus data, and controlled native speaker judgment tasks strongly supports the hypothesis that this conflict led to the construction's avoidance and eventual disappearance.

A critical part of our proposal is that the functional replaceability of -yā: kar by other constructions allowed this gap to become entrenched without any loss of expressivity. To further validate this mechanism of change, a crucial direction for future work is to conduct a comparative diachronic frequency analysis of the alternative -tā rah- construction. Tracking the frequency of -tā rah- in its habitual sense across the same historical corpus would allow us to test the hypothesis that its rise coincided with the decline of the perfective -yā: kar form. This would provide powerful quantitative evidence for the role of constructional competition and functional replacement in the creation of this paradigm gap.

\nocite{*}
\bibliography{template}
\end{document}